\documentclass{article}

\usepackage{proceedings}
\usepackage{times}
\usepackage{epsfig}
\usepackage{amssymb}
\usepackage{named}

\def \snake	 {\smash{|\!\!\!\sim}}

\def\definedas{{\:\:\stackrel{{\it def}}{=}\:\:}}
\def\DW{{W\hspace{-1mm}\downarrow}}
\def\holds{{\it holds}}
\def\shrink(#1){}

\newcommand{\Omit} [1] {} 

\newtheorem{definition}{Definition}[section]

\newtheorem{corollary}{Corollary}[section]
\newtheorem{proposition}{Proposition}[section]
\newtheorem{example}{Example}[section]

\title{Compilation of Propositional Weighted Bases}

\author{Adnan Darwiche\\
Computer Science Department \\
University of California, Los Angeles \\
CA 90095, USA \\
{\small e-mail: {\tt darwiche@cs.ucla.edu}} \And
Pierre Marquis\\
CRIL-CNRS \\
 Universit\'e d'Artois\\
F-62307, Lens Cedex, France \\
 {\small e-mail: {\tt
marquis@cril.univ-artois.fr}} }

\begin{document}

\maketitle
\bibliographystyle{named}

\begin{abstract}
In this paper, we investigate the extent to which knowledge
compilation can be used to improve inference from propositional
weighted bases. We present a general notion of compilation of a
weighted base that is parametrized by any equivalence--preserving
compilation function. Both negative and positive results are
presented. On the one hand, complexity results are identified,
showing that the inference problem from a compiled weighted base
is as difficult as in the general case,
when the prime implicates, Horn cover
or renamable Horn cover classes are targeted.
On the other hand, we show that the inference problem becomes tractable whenever
$DNNF$-compilations are used and clausal queries are considered.
Moreover, we show that the set of all preferred models of a
$DNNF$-compilation of a weighted base can be computed in
time polynomial in the output size. Finally, we sketch
how our results can be used in model-based diagnosis in order to compute
the most probable diagnoses of a system.
\end{abstract}

\section{INTRODUCTION}\label{introduction}

{\em Penalty logic} is a logical framework developed by Pinkas
\cite{Pinkas91,Pinkas95} and by Dupin de St Cyr, Lang
and Schiex \cite{Dupinetal94}, which enables the representation of
propositional weighted bases. A {\em weighted base} is a finite set
$$W = \{\langle \phi_1, k_1\rangle, \ldots, \langle \phi_n, k_n\rangle\}.$$
Each $\phi_i$ is a propositional formula, and $k_i$ is its corresponding weight, i.e.,
the price to be paid if the formula is violated. In penalty logic, weights are
positive integers\footnote{Floating numbers can also be used; what is important
is the fact that sum is a total function over the set of (totally ordered) numbers under consideration,
and that it can be computed in polynomial time.}
or $+\infty$ and they are {\em additively} aggregated.

A weighted base can be considered as a compact, implicit encoding
of a total pre-ordering over a set $\Omega$ of propositional
worlds. Indeed, given a weighted base $W$, the weight of each
world $\omega$ can be defined as follows:
\[ K_W(\omega) \definedas
\sum_{\langle \phi_i, k_i\rangle \in W, \: \omega \models \neg
\phi_i} k_i. \] That is, the weight of a world is the sum of all
weights associated with sentences violated by the world.
One can extend the function \(K_W\) to arbitrary sentences
\(\alpha\):
\[ K_W(\alpha) \definedas \min_{\omega \models \alpha} K_W(\omega).\]
Finally, $min_W(\Omega)$ denotes the most preferred worlds in
$\Omega$, those having minimal weight:
 \[ min_W(\Omega) \definedas \{\omega \: | \: \omega \in \Omega, \forall \omega^\prime \in \Omega \: 
K_W(\omega) \leq K_W(\omega^\prime)\}.\]
 The weight of base \(W\), denoted \(K(W)\), is the weight of some
 world in \(min_W(\Omega)\).
Obviously enough, we have $K(W) = K_W(true)$, and
$\omega \in min_W(\Omega)$
if and only if $K_W(\omega) = K(W)$.

\begin{example}
Let $W = \{\langle a \wedge b, 2\rangle, \langle \neg b,
1\rangle\}$ be a weighted base. Let us consider the following four
worlds over the variables appearing in \(W\), $Var(W)$:
\begin{itemize}
\item $\omega_1 = (a, b)$
\item $\omega_2 = (a, \neg b)$
\item $\omega_3 = (\neg a, b)$
\item $\omega_4 = (\neg a, \neg b)$
\end{itemize}
We then have $K_W(\omega_1) = 1$, $K_W(\omega_2) = 2$,
$K_W(\omega_3) = 3$, and $K_W(\omega_4) = 2$. Accordingly, we have
$K(W) = 1$ and $min_W(\Omega) = \{\omega_1\}$.
\end{example}

All formulas $\phi_i$ associated with finite weights in a weighted
base are called {\em soft constraints}, while those associated
with the weight $+\infty$ are called {\em hard constraints}.


Penalty logic has some valuable connections with possibilistic
logic, as well as with Dempster--Shafer theory (see
\cite{Dupinetal94} for details). It is also closely connected to
the optimization problem {\sc weighted-max-sat} considered in
operations research. Accordingly, several proposals for the use of
weighted bases can be found in the AI literature.

One of them concerns the compact representation of
preferences in a decision making setting. Indeed,
in some decision making problems, models (and formulas)
can be used to encode decisions.  Accordingly, the weight of a model
represents the disutility of a decision, and a weighted base
can be viewed as an implicit representation of the set of all decisions
of an agent, totally ordered w.r.t. their (dis)utility. Lafage and Lang
\cite{LafageLang00} take advantage of such an encoding for group
decision making. A key issue here from a computational point of view
is the
problem consisting in computing (one or all) element(s) from $min_W(\Omega)$.

Another suggested use of penalty logic concerns inference from
inconsistent belief bases. Based on the preference information
given by $K_W$, several inference relations from a weighted base
$W$ can be defined. Among them is skeptical inference given by
$\alpha \snake_W \beta$ if and only if every world $\omega$ that
is of minimal weight among the models of $\alpha$ is a
model of $\beta$. In this framework, propositional formulas
represent pieces of (explicit) belief. The inference relation
$\snake_W$ is interesting for at least two reasons. On the one
hand, it is a {\em comparative} inference relation, i.e., a
rational inference relation satisfying supraclassicality
\cite{Dupinetal94}. On the other hand, weighted bases can be used
to encode some well-known forms of inference from stratified
belief bases $B = (B_1, \ldots, B_k)$
\cite{PinkasLoui92,Benferhatetal93,Benferhatetal95}. Especially,
the so-called skeptical {\em lexicographic inference} $B~
\snake_{lex}$ can be recovered as a specific case of $true~
\snake_{W_B}$ for some weighted base $W_B$.

\begin{example}
Let $B = (B_1, B_2)$ be a belief base interpreted under
lexicographic inference, where $B_1 = \{a \vee b \vee c\}$ (the
most reliable stratum) and $B_2 = \{\neg a \wedge c, \neg b \wedge
c, \neg c\}$. W can associate with $B$ the weighted base 
$$W_B =
\{\langle a \vee b \vee c, 4\rangle, \langle \neg a \wedge c,
1\rangle, \langle \neg b \wedge c, 1\rangle, \langle \neg c,
1\rangle\}.$$ 
\noindent The unique most preferred world for $W_B$ is $(\neg
a, \neg b, c)$ that is also the only lexicographically-preferred
model of $B$.
\end{example}

Weighted bases enable more flexibility than stratified belief
bases (e.g., violating two formulas of weight $5$ is worse than
violating a single formula of weight $9$, but this cannot be
achieved through a simple stratification)\footnote{Since
lexicographic inference also includes inference from consistent
sub--bases that are maximal w.r.t. cardinality as a subcase (to
achieve it, just put every formula of the belief base into a
single stratum), the latter can also be recovered as a specific
case of inference from a weighted base.}.

The {\em inference} problem from a weighted base \(W\) consists in
determining whether $true \snake_W \beta$ holds given $W$ and
$\beta$. Up to now, weighted bases have been investigated from a
theoretical point of view, only. Despite their potentialities, we
are not aware of any industrial application of weighted bases.
There is a simple (but partial) explanation of this fact:
inference (and preferred model enumeration) from weighted bases
are intractable. Actually, the inference problem is known as
$\Delta_2^p$-complete \cite{Dupin96} (even in the restricted case
where queries are literals). Furthermore, it is not hard to show
that computing a preferred world from $min_W(\Omega)$ is {\sf
F}$\Delta_2^p$-complete. 
This implies that any
of the two problems is very likely to require an unbounded
polynomial number of calls to an {\sf NP} oracle to be solved in
polynomial time on a deterministic Turing machine.

In this paper, we investigate the extent to which {\em knowledge compilation}
\cite{CadoliDonini97} can be used to improve inference from weighted bases.
The key idea of compilation is pre-processing of the fixed part of the inference
problem. Several knowledge compilation functions dedicated to the clausal
entailment problem have been pointed out so far
(e.g., \cite{ReiterdeKleer87,delVal94,DechterRish94,Marquis95,Dalal96,Schrag96,SelmanKautz96,Boufkhadetal97,Darwiche99}).
The input formula is turned into a compiled one during an off-line compilation phase and
the compiled form is used to answer the queries on-line.
Assuming that the formula does not often change
and that answering queries from the compiled form is computationally easier than
answering them from the input formula, the compilation time can be balanced over
a sufficient number of queries. Thus, when queries are CNF formulas,
the complexity of classical inference falls from
{\sf coNP}-complete to {\sf P}. While none of the techniques listed above
can ensure the objective of enhancing inference to be reached in the worst case
(because the size of the compiled form can be exponentially larger
than the size of the original knowledge base -- see \cite{SelmanKautz96,CadoliDonini97}),
experiments have shown such approaches valuable
in many practical situations \cite{Schrag96,Boufkhadetal97,darwicheD125}.

In the following, we show how compilation functions for clausal
entailment from classical formulas can be extended to clausal
inference from weighted bases. Any equivalence--preserving
knowledge compilation function can be considered in our framework.
Unfortunately, for many target classes for such functions,
including the prime implicates, Horn cover and renamable Horn cover classes,
we show that the inference problem from
a compiled base remains $\Delta_2^p$-complete, even for very simple queries (literals).
Accordingly, in this situation,
there is no guarantee that compiling a weighted base using any of
the corresponding compilation functions may help. Then we focus on
$DNNF$-compilations as introduced in
\cite{Darwiche99,darwicheJANCL}. This case is much more favourable
since the clausal inference problem becomes tractable. We also
show that the preferred models of a
$DNNF$-compilation of a weighted base can be enumerated in output
polynomial time. Finally, we sketch how our results can be used in
the model--based diagnosis framework in order to compute the most
probable diagnoses of a system.

\section{FORMAL PRELIMINARIES}\label{formal}

In the following, we consider a propositional language
$PROP_{PS}$ defined inductively from
a finite set $PS$ of propositional symbols, the boolean
constants $true$ and $false$ and the
connectives $\neg$, $\wedge$, $\vee$ in the usual way.
$L_{PS}$ is the set of literals built up from $PS$.
For every formula $\phi$ from $PROP_{PS}$,
$Var(\phi)$ denotes the symbols of $PS$ occurring in
$\phi$. 
As mentioned before, if $W = \{\langle \phi_1, k_1\rangle, \ldots, \langle \phi_n, k_n\rangle\}$
is a weighted base, then $Var(W) = \bigcup_{i=1}^n Var(\phi_i)$.

Formulas are interpreted in a classical way. As evoked before,
$\Omega \: (= 2^{PS})$ denotes the set of all interpretations
built up from $PS$. Every interpretation (world) $\omega \in
\Omega$ is represented as a tuple of literals. $Mod(\phi)$ is the
set of all models of $\phi$.

As usual, every finite set of formulas is considered as the conjunctive
formula whose conjuncts are the elements of the set.
A CNF formula is a (finite) conjunction of clauses, where a clause
is a (finite) disjunction of literals. A formula $\phi$ is Blake
if and only if it is a CNF formula where each prime
implicate\footnote{A prime implicate of a formula $\phi$ is a
logically strongest clause entailed by $\phi$.} of $\phi$ appears
as a conjunct (one representative per equivalence class). A
formula is Horn CNF if and only if it is a CNF formula s.t. every
clause in it contains at most one positive literal. A formula
$\phi$ is renamable Horn CNF if and only if $\sigma(\phi)$ is a
Horn CNF formula, where $\sigma$ is a substitution from $L_{PS}$
to $L_{PS}$ s.t. $\sigma(l) = l$ for every literal $l$ of $L_{PS}$
except those of a set $L$, and for every literal $l$ of $L$,
$\sigma(l) = \neg l$ and $\sigma(\neg l) = l$.

We assume the reader familiar with
the complexity classes {\sf P}, {\sf NP}, {\sf coNP}
and $\Delta_2^p$ of the polynomial hierarchy.
{\sf F}$\Delta_2^p$ denotes the class of function problems associated
to $\Delta_2^p$ (see \cite{Papadimitriou94} for details).

\section{COMPILING WEIGHTED BASES}

In this section, we first show how knowledge compilation techniques
for improving clausal entailment can be used in order to compile
weighted bases. Then, we present some complexity results showing that
compiling a weighted base is not always a good idea, since the complexity
of inference from a compiled base does not necessarily decrease.
We specifically focus on prime implicates \cite{ReiterdeKleer87} and
Horn covers and renamable Horn covers compilations \cite{Boufkhadetal97}.

\subsection{A FRAMEWORK FOR WEIGHTED BASES COMPILATION}

Let $W = \{\langle \phi_1, k_1\rangle, \ldots, \langle \phi_n,
k_n\rangle\}$ be a weighted base. In the case where
$\bigwedge_{i=1}^n \phi_i$ is consistent, then $K(W) = 0$ and
$min_W(\Omega)$ is the set of all models of $\bigwedge_{i=1}^n
\phi_i$. Accordingly, in this situation, inference $\snake_W$ is
classical entailment, so it is possible to directly use any
knowledge compilation function and compiling $W$ comes down to
compile $\bigwedge_{i=1}^n \phi_i$. However, this situation is
very specific and out of the ordinary when weighted bases are
considered (otherwise, weights would be useless). A difficulty is
that, in the situation where $\bigwedge_{i=1}^n \phi_i$ is
inconsistent, we cannot compile directly this formula using any
equivalence--preserving knowledge compilation function (otherwise,
trivialization would not be avoided). Indeed, in this situation,
$\snake_W$ is not classical entailment any longer, so a more
sophisticated approach is needed.

In order to compile weighted bases, it is helpful to consider
weighted bases in normal form:

\begin{definition}[Weighted bases in normal form]~\\
A belief base $W = \{\langle \phi_1, k_1\rangle, \ldots, \langle
\phi_n, k_n\rangle\}$ is {\em in normal form} if and only if for
every $i \in 1 \ldots n$, either $k_i = +\infty$ or $\phi_i$ is a
propositional symbol.
\end{definition}

Every weighted base can be turned into a query--equivalent base in
normal form.

\begin{definition}[$V$-equivalence of
weighted bases]~\\
Let $W_1$ and $W_2$ be two weighted bases
and let \(V \subseteq PS\). $W_1$ and $W_2$ are
$V$-{\em equivalent} if and only if for every pair of sentences
\(\alpha\) and \(\beta\) in  $PROP_{V}$, we have \(\alpha
\snake_{W_1} \beta\) precisely when \(\alpha \snake_{W_2} \beta\).
\end{definition}

Accordingly, two $V$--equivalent weighted bases must agree on
queries built up from the symbols in $V$. Note that a stronger
notion of equivalence can be defined by requiring that both bases
induce the same weight function, i.e., $K_{W_1} = K_{W_2}$
\cite{Dupinetal94}. Finally, note that if \(K_{W_1}\) and
\(K_{W_2}\) agree on the sentences in \(PROP_V\), then \(W_1\) and
\(W_2\) must be \(V\)--equivalent.

\begin{proposition}\label{normal}~\\
Let $W = \{\langle \phi_1, k_1\rangle, \ldots, \langle \phi_n,
k_n\rangle\}$ be a weighted base. If
\begin{eqnarray*}
H & \definedas & \{\langle \phi_i, +\infty\rangle \ | \ \langle
\phi_i, +\infty\rangle \in W\}, \\
S & \definedas & \{\langle \holds_i \ \Rightarrow \ \phi_i,
+\infty\rangle, \langle \holds_i, k_i\rangle \ | \\
& &  \:\:\:\:\:\: \langle \phi_i, k_i\rangle \in W and \ k_i \neq
+\infty \:\: \},
\end{eqnarray*}
where $\{\holds_1, \ldots, \holds_n\} \subseteq PS \setminus
Var(W)$, then the weighted base \(\DW \definedas H \cup S\) is in
normal form.

Moreover, \(K_W\) and \(K_{\DW}\) agree on all weights of
sentences in \(PROP_{Var(W)}\) and, hence, \(\DW\) is
$Var(W)$--equivalent to $W$.
\end{proposition}

We will call $\DW$ the normal form of $W$. Intuitively, the
variable $\holds_i$ is guaranteed to be false in any world that
violates the sentence $\phi_i$ and, hence, that world is
guaranteed to incur the penalty \(k_i\).

\noindent {\bf Example 1.1} (Continued) The weighted base \\
$$\DW = \{\langle \holds_1 \Rightarrow (a \wedge b), +\infty\rangle,$$
$$\langle \holds_2 \Rightarrow \neg b, +\infty\rangle, \langle \holds_1, 2\rangle, \langle \holds_2,
1\rangle\}$$ 
\noindent is a normal form of $W$ as given by
Proposition~\ref{normal}. The normalized weighted base \(\DW\)
induces the following weight function:

\begin{center}
\begin{tabular}{l|l}
World & $K_\DW$ \\ \hline
$a,b,\holds_1,\holds_2$ & $+\infty$ \\
$a,b,\holds_1,\neg \holds_2$ & 1 \\
$a,b,\neg \holds_1,\holds_2$ & $+\infty$ \\
$a,b,\neg \holds_1,\neg \holds_2$ & 3 \\ \hline
$a,\neg b,\holds_1,\holds_2$ & $+\infty$ \\
$a,\neg b,\holds_1,\neg \holds_2$ & $+\infty$ \\
$a,\neg b,\neg \holds_1,\holds_2$ & 2 \\
$a,\neg b,\neg \holds_1,\neg \holds_2$ & 3 \\ \hline
$\neg a,b,\holds_1,\holds_2$ & $+\infty$\\
$\neg a,b,\holds_1,\neg \holds_2$ & $+\infty$ \\
$\neg a,b,\neg \holds_1,\holds_2$ & $+\infty$\\
$\neg a,b,\neg \holds_1,\neg \holds_2$ & 3 \\ \hline
$\neg a,\neg b,\holds_1,\holds_2$ & $+\infty$ \\
$\neg a,\neg b,\holds_1,\neg \holds_2$ & $+\infty$ \\
$\neg a,\neg b,\neg \holds_1,\holds_2$ & 2 \\
$\neg a,\neg b,\neg \holds_1,\neg \holds_2$ & 3 \\
\end{tabular}
\end{center}

We have $K(\DW) = 1$ and
$$min_{\DW}(\Omega) = \{(a, b, \holds_1, \neg \holds_2)\}.$$
Moreover, \(K_W\) and \(K_\DW\) agree on all
sentences constructed from the variables in \(\{a,b\}\).

We now discuss the compilation of a weighted base in normal form.
The basic idea is to combine all hard constraints in the base into
a single constraint, which preserves the weight function induced
by the base. We then compile that single hard constraint using an
equivalence--preserving compilation function $COMP$, that is, a
function which maps each sentence \(\alpha\) into its compiled
form $COMP(\alpha)$. From here on, we will use \(\widehat{W}\) to
denote the conjunction of all
sentences in the weighted base \(W\) that have \(+\infty\)
weights:
\[ \widehat{W} \definedas
\bigwedge_{\langle \phi_i, +\infty\rangle \in W} \phi_i. \]

\begin{definition}[Compilation of a weighted base]\label{defcompilwb}~\\
Let $W = \{\langle \phi_1, k_1\rangle, \ldots, \langle \phi_n,
k_n\rangle\}$ be a weighted base. Let $COMP$ be any
equivalence--preserving knowledge compilation function. The
$COMP$-compilation of $W$ is the weighted base
$$\DW_{COMP} \definedas \{\langle COMP(\widehat{\DW}),
+\infty\rangle\} \cup $$
$$\{\langle \holds_i, k_i\rangle \ | \ \langle \holds_i, k_i\rangle \in \DW
\ and \ k_i \neq +\infty\}.$$
\end{definition}

That is, to compile a weighted base \(W\), we perform three steps.
First, we compute a normal form \(\DW\) according to
Proposition~\ref{normal}, which is guaranteed to be
\(V\)-equivalent to \(W\), where \(V\) are the variables in \(W\).
Next, we combine all of the hard constraints of \(\DW\) into a
single hard constraint \(\widehat{\DW}\). Finally, we compile
\(\widehat{\DW}\) using the function \(COMP\).

\noindent {\bf Example 1.1} (Continued) We have 
$$\widehat{\DW} = (\neg \holds_1 \vee (a \wedge b)) \wedge (\neg \holds_2 \vee \neg b).$$
Accordingly, the Blake-compilation of $W$ is 
$$\{\langle (\neg
\holds_1 \vee a) \wedge (\neg \holds_1 \vee b) \wedge (\neg
\holds_2 \vee \neg b)$$
$$\wedge (\neg \holds_1 \vee \neg \holds_2),
+\infty\rangle, \langle \holds_1, 2\rangle, \langle \holds_2,
1\rangle\}.$$

Given Proposition \ref{normal}, and since $COMP$ is
equivalence--preserving, we have:

\begin{corollary}
Let $W = \{\langle \phi_1, k_1\rangle, \ldots, \langle \phi_n,
k_n\rangle\}$ be a weighted base.  Let $COMP$ be any
equivalence--preserving knowledge compilation function.
$K_{\DW_{COMP}}$ and $K_W$ agree on the sentences in
\(PROP_{Var(W)}\). Moreover, $W$ and $\DW_{COMP}$ are
$Var(W)$-equivalent.
\end{corollary}

\subsection{SOME COMPLEXITY RESULTS}

In the following, the next tractable classes of formulas, that are
target classes for some existing equivalence--preserving
compilation functions $COMP$, are considered:
\begin{itemize}
\item The {\em Blake} class is the set of formulas given in prime implicates normal form,
\Omit{\footnote{Formulas
from this class can be recognized in polynomial time since a
given CNF formula $\Sigma$ is Blake if and only if
every possible resolvent between two clauses of $\Sigma$ is entailed by a clause of $\Sigma$,
and no clause is entailed by another clause of $\Sigma$.},}
\item The {\em Horn cover} class is the set of disjunctions of Horn CNF formulas,
\item The {\em renamable Horn cover} class
(r. Horn cover for short)
is the set of the disjunctions of
renamable Horn CNF formulas.
\end{itemize}

The Blake class is the target class of the compilation function $COMP_{Blake}$ described
in \cite{ReiterdeKleer87}. The Horn cover class and the renamable Horn cover class
are target classes for the tractable covers
compilation functions given in \cite{Boufkhadetal97}. We shall note respectively
$COMP_{Horn~cover}$ and
$COMP_{r.~Horn~cover}$ the corresponding compilation functions.

Accordingly, a Blake (resp. Horn cover, r. Horn cover) compiled weighted base
$W$ is defined as a weighted base in normal form whose unique hard constraint
belongs to the Blake (resp. Horn cover, r. Horn cover) class.

In the next section, we will also focus on the {\em DNNF} class. We consider it
separately because --- unlike the other classes --- it will lead to
render tractable clausal inference from
compilations.

Of course, all these compilation functions $COMP$ are subject to the limitation evoked above: in the
worst case, the size of the compiled form $COMP(\Sigma)$ is exponential in the size of $\Sigma$.
Nevertheless, there is some empirical evidence that some of these approaches can prove computationally
valuable for many instances of the clausal entailment problem (see e.g., the experimental
results given in \cite{Schrag96,Boufkhadetal97,darwicheD125}).

As evoked previously, knowledge compilation can prove helpful only if inference from
the compiled form is computationally easier than direct inference. Accordingly, it is
important to identify the complexity of inference from a compiled weighted base
if we want to draw some conclusions about the usefulness of knowledge compilation
in this context. Formally, we are going to consider the following decision problems:

\begin{definition}[{\sc formula $\snake_W$}]{~}\\
{\sc formula $\snake_W$} is the following decision problem:\\
\vspace{-4mm}
\begin{itemize}
\item {\bf Input:} A weighted base $W$ and a formula $\beta$ from $PROP_{PS}$.
\item {\bf Query:} Does $true \snake_W \beta$ hold?
\end{itemize}
\end{definition}

{\sc clause $\snake_W$} (resp. {\sc literal $\snake_W$}) is the restriction
of {\sc formula $\snake_W$} to the case where $\beta$ is required to be a CNF formula
(resp. a term).

When no restriction is put on $W$,
{\sc formula $\snake_W$} is known as $\Delta_2^p$-complete \cite{Dupin96},
even in the restricted {\sc literal $\snake_W$} case. Now, what if $W$ is a
{\em compiled} weighted base? We have identified the following results:

\begin{proposition}[Inference from compiled weighted bases]\label{comp}
The complexity of {\sc clause $\snake_W$}
and of its restrictions to literal inference when $W$ is a
$Blake$ (resp. $Horn~cover$, $r.~Horn~cover$) compiled
weighted base
is reported in Table \ref{compiled}.
\end{proposition}
\begin{table}
\begin{center}
\begin{tabular}{|c|c|}\hline
{\bf $COMP$} & {\bf {\sc clause  / literal $\snake_W$}} \\
\hline \hline
$Blake$ & $\Delta_2^p$-complete\\ \hline
$Horn~cover$ & $\Delta_2^p$-complete\\ \hline
$r~Horn~cover$ & $\Delta_2^p$-complete\\ \hline
\end{tabular}
\caption{Complexity of clausal inference from compiled weighted bases.}\label{compiled}
\end{center}
\end{table}

Hardness results can be easily derived from results given in \cite{CosteMarquis00}
due to the fact that (skeptical) lexicographic inference $\snake_{lex}$ from
a stratified belief base can be easily encoded as inference from a weighted base.
Indeed, if $m$ is the maximum number of formulas belonging to any
stratum $B_i$ of $B = (B_1, \ldots, B_k)$,
then let $W_B = \{\langle \phi, (m+1)^{k-i}\rangle \ | \ \phi \in B_i\}$
(see Example 1.2 for an illustration).
It is not hard to prove that $B~ \snake_{lex} \beta$ if and only if
$true~ \snake_{W_B} \beta$.

The complexity results reported in Table \ref{compiled} do not
give good news: there is no guarantee that compiling a belief base using
the $Blake$ (or the $Horn~cover$ or the $r.~Horn~cover$) compilation function
leads to improve inference since its complexity from the
corresponding compiled bases is just as hard as the complexity
of $\snake_W$ in the general case.

Fortunately, it is not the case that such negative results hold for any
compilation function. As we will see in the next section,
$DNNF$-compilations of weighted bases exhibit a much better behaviour.

\section{COMPILING WEIGHTED BASES USING DNNF}

In this section, we focus on $DNNF$-compilations of weighted bases.
After a brief recall of what $DNNF$-compilation is, we show that
$DNNF$-compilations support two important computational
tasks in polynomial time, especially preferred model enumeration and clausal
inference.

\subsection{A GLIMPSE AT THE $DNNF$ LANGUAGE}

The $DNNF$ language is the set of sentences, defined as follows \cite{Darwiche99}:

\begin{definition}[DNNF]~
Let $PS$ be a finite set of propositional variables.
A sentence in $DNNF$ is a rooted, directed acyclic graph (DAG) where each leaf node is
labeled with \(true\), \(false\), \(x\) or \(\neg x\), \(x \in
PS\); each internal node is labeled with \(\wedge\) or
\(\vee\) and can have arbitrarily many children.
Moreover, the {\em decomposability} property is satisfied: for each conjunction \(C\) in
the sentence, the conjuncts of \(C\) do not share variables.
\end{definition}

\begin{figure}[tb]
\begin{center}
\mbox{\epsffile{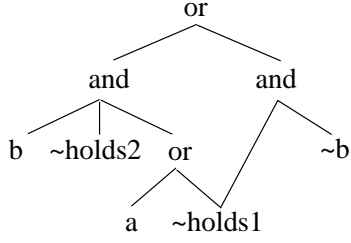}}
\end{center}
\caption{A sentence in $DNNF$.\label{fig1}}
\end{figure}

\shrink({Figure \ref{fig1} depicts some $DNNF(\widehat{\DW})$,
where $W$ is the weighted base given in Example 1.1. Note here
that \(\DW\) is the normal form of \(W\), and \(\widehat{\DW}\) is
the conjunction of all sentence in \(\DW\) that have weight
\(+\infty\).})

Figure \ref{fig1} depicts a $DNNF$ of the hard constraint
$\widehat{\DW}$, where $W$ is the weighted base given in Example
1.1. Note here that \(\DW\) is 
the normal form constructed from
\(W\) according to Proposition~\ref{normal}, and \(\widehat{\DW}\)
is the conjunction of all hard constraints in \(\DW\).

An interesting subset of $DNNF$ is the set of {\em smooth} $DNNF$
sentences \cite{darwicheJANCL}:

\begin{definition}[Smooth DNNF]~
A $DNNF$ sentence satisfies the {\em smoothness} property
if and only if for each disjunction \(C\) in the sentence, each
disjunct of \(C\) mentions the same variables.
\end{definition}

Interestingly, every $DNNF$ sentence can be turned into an
equivalent, smooth one in polynomial time \cite{darwicheJANCL}.

\begin{figure}[tb]
\begin{center}
\mbox{\epsffile{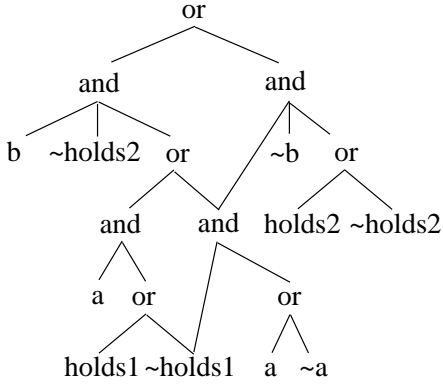}}
\end{center}
\caption{A sentence in smooth $DNNF$.\label{fig2}}
\end{figure}

For instance, Figure~\ref{fig2} depicts a smooth $DNNF$ which is
equivalent to the $DNNF$ in Figure~\ref{fig1}. Note that for
readibility reasons some leaf nodes are duplicated in the figure.

Among the various tasks that can be achieved in a tractable way from a
$DNNF$ sentence are {\em conditioning}, {\em clausal
entailment}, {\em forgetting} and {\em model enumeration} (given
that the $DNNF$ is smooth) \cite{darwicheJANCL,DarwicheMarquis01}.

\subsection{TRACTABLE QUERIES}

Given a weighted base \(W\), and given a $DNNF$--compilation of
\(W\), we now show how the compilation can be used to represent
the preferred models of \(W\) as a $DNNF$ in polynomial time.

\begin{definition}[Minimization of a weighted base]~\\
A {\em minimization} of a weighted base $W$ is a propositional
formula $\Delta$ where the models of \(\Delta\) are
$\min_W(\Omega)$.
\end{definition}

Note that this notion generalizes
the notion of minimization of a propositional formula $\phi$
reported in \cite{Darwiche99}, for which the
preferred models are those containing a maximal number of variables
assigned to true. Such a minimization can be easily achieved in a weighted base
setting by considering the base $\{\langle \phi, +\infty\rangle\} \cup
\bigcup_{x \in Var(\phi)} \{\langle x, 1\rangle\}$.

\begin{definition}[Minimization of $DNNF$-compilation]
Let \(W\) be a $DNNF$-compilation of a weighted base. Let $\langle
\alpha,+\infty\rangle$ be the single hard constraint in \(W\),
where \(\alpha\) is a $DNNF$ sentence. Suppose that \(\alpha\) is
also smooth.
\begin{itemize}
\item We define $k(\alpha)$ inductively as follows:
\begin{itemize}
\item \(k(true) \definedas 0\) and \(k(false) \definedas +\infty\).
\item If \(\alpha\) is a literal, then:
\begin{itemize}
\item If $\alpha=\neg\holds_i$, then $k(\alpha) \definedas k_i$, where
\(\langle \holds_i,k_i\rangle \in W\).
\item Otherwise, $k(\alpha) \definedas 0$.
\end{itemize}
\item $k(\alpha = \bigvee_i \alpha_i) \definedas min_i k(\alpha_i)$.
\item $k(\alpha = \bigwedge_i \alpha_i) \definedas \sum_i k(\alpha_i)$.
\end{itemize}
\item We define $min(\alpha)$ inductively as follows:
\begin{itemize}
\item If $\alpha$ is a literal or a boolean
constant, then $min(\alpha) \definedas \alpha$.
\item $min(\alpha = \bigvee_i \alpha_i) \definedas \bigvee_{k(\alpha_i) = k(\alpha)} min(\alpha_i)$.
\item $min(\alpha = \bigwedge_i \alpha_i) \definedas \bigwedge_i min(\alpha_i)$.
\end{itemize}
\end{itemize}
\end{definition}

We have the following result:

\begin{proposition}
Let \(W\) be
a $DNNF$-compilation of a weighted base. Let $\langle
\alpha,+\infty\rangle$ be the single hard constraint in \(W\),
where \(\alpha\) is a smooth $DNNF$ sentence. Then $min(\alpha)$
is a smooth $DNNF$ and is a minimization of $W$.
\end{proposition}

\begin{figure}[tb]
\begin{center}
\mbox{\epsffile{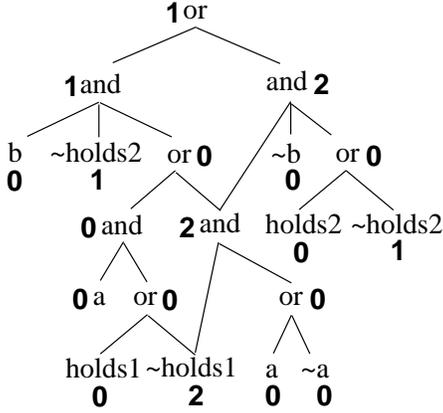}}
\end{center}
\caption{Weights on a smooth $DNNF$ sentence.\label{fig3}}
\end{figure}

\begin{figure}[tb]
\begin{center}
\mbox{\epsffile{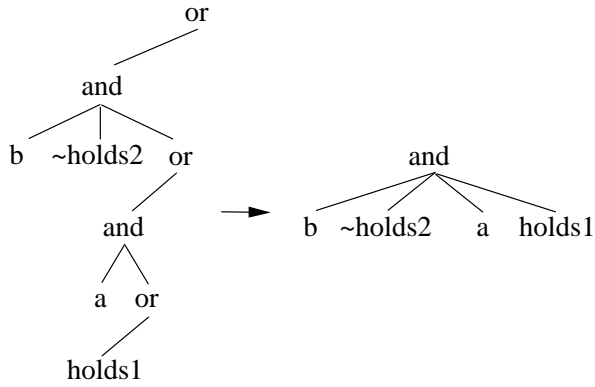}}
\end{center}
\caption{Minimization of a smooth $DNNF$ sentence.\label{fig4}}
\end{figure}

Figure \ref{fig3} depicts the weight $k(\alpha)$ of every
subformula $\alpha$ of the smooth $DNNF$ sentence given in Figure
\ref{fig2}. Figure~\ref{fig4} (left) depicts the minimization of
the $DNNF$ in Figure~\ref{fig3}. Figure~\ref{fig4} (right) depicts
a simplification of this minimized $DNNF$ which has a single
model.

Since $min(\alpha)$ can be computed in time polynomial in the size
of $DNNF$ $\alpha$, and since clausal entailment can be done in
time linear in the size of \(\alpha\) \cite{Darwiche99}, we have:

\begin{corollary}
The clausal inference problem {\sc clause $\snake_W$} from
$DNNF$-compilations of weighted bases is in {\sf P}.
\end{corollary}

Since model enumeration can be done in output polynomial time from
a smooth $DNNF$, we also have:

\begin{corollary}
The preferred model enumeration problem from
$DNNF$-compilations of weighted bases can be solved in output
polynomial time.
\end{corollary}

\section{APPLICATION TO MODEL-BASED DIAGNOSIS}

We now briefly sketch how the previous results can be used to
compute the set of most probable diagnoses of a system in time
polynomial in the size of system description and the output size.
The following results generalize those given in
\cite{Darwiche98,darwicheJANCL} to the case where the probability
of failure of components is available.

We first need to briefly recall what a consistency-based
diagnosis of a system is \cite{Reiter87}:

\begin{definition}[Consistency-based diagnosis]~\\
\vspace{-5mm}
\begin{itemize}
\item A {\em diagnostic problem} ${\cal P} = \langle SD, OK, OBS\rangle$
is a triple consisting of:
\begin{itemize}
\item a formula $SD$ from $PROP_{PS}$, the system description.
\item a finite set $OK = \{ok_1, \ldots, ok_n\}$ of propositional symbols.
``$ok_i$ is true'' means that the component $i$ of the system to be
diagnosed is not faulty.
\item $OBS$ is a term, typically gathering the
inputs and the outputs of the system.
\end{itemize}
\item A {\em consistency-based} diagnosis $\Delta$ for ${\cal P}$ is
a complete $OK$-term (i.e., a conjunction of literals built up from $OK$
in which every $ok_i$ occurs either positively or negatively)
s.t. $\Delta \wedge SD \wedge OBS$ is consistent.
\end{itemize}
\end{definition}

Because a system can have a number of diagnoses that is exponential
in the number of its components, preference criteria are usually
used to limit the number of candidates. The most current ones consist
in keeping the diagnoses containing as few negative $OK$-literals
as possible (w.r.t. set inclusion or cardinality).

When the {\it a priori} probability of failure of components
is available (and such probabilities are considered
independent), the most probable diagnoses for ${\cal P}$ can also
be preferred. Such a notion of preferred diagnosis generalizes
the one based on minimality w.r.t. cardinality (the latter corresponds
to the case where the probability of failure of components is uniform
and $< \frac{1}{2}$).

Interestingly, the most probable diagnoses for ${\cal P}$ can be enumerated in
output polynomial
time as soon as a smooth
$DNNF$-compilation ${\cal P}_{DNNF}$ corresponding
to ${\cal P}$ has been derived first.

\begin{definition}[Compilation of a diagnostic problem]~\\
Let ${\cal P}$ be a diagnostic problem for which the {\it a priori}
probability of failure $p_i$ of any component $i$ is available.\\
$${\cal P}_{DNNF} \definedas \{\langle DNNF(SD\ | \ OBS),
+\infty\rangle\} \cup $$
$$\{\langle ok_i, log \ p_i\rangle \ | \ ok_i
\in OK\}$$ 
is the smooth
$DNNF$-compilation associated
with ${\cal P}$.
\end{definition}

In this definition, $SD \ | \ OBS$ denotes the conditioning of $SD$ by
the term $OBS$, i.e., the formula obtained by replacing in $SD$ every variable $x$
by $true$ (resp. $false$) if $x$ (resp. $\neg x$) is a positive (resp. negative) literal
of $OBS$.

The $log$ transformation performed here enables to compute the
$log$ of the probability of a diagnosis $\Delta$ as $\sum_{\neg ok_i \in \Delta} log \ p_i$.
Because $log$ is strictly nondecreasing, the induced preference ordering between
diagnoses is preserved.

\begin{proposition}~\\
\vspace{-5mm}
\begin{itemize}
\item $K({\cal P}_{DNNF})$ is the $log$ of the probability
of any most probable diagnosis for ${\cal P}$.
\item The most probable diagnoses for ${\cal P}$ are the
the models of $Forget(min(DNNF(SD \ | \ OBS)), PS \setminus OK)$\footnote{For every
formula $\phi$ and every set of variables $X$, $Forget(\phi, X)$ denotes the
logically strongest consequence of $\phi$ that is independent from $X$,
i.e., that can be turned into an equivalent formula in which no
variable from $X$ occurs.}.
\end{itemize}
\end{proposition}

An important point is that ${\cal P}_{DNNF}$ does not have to be re-compiled
each time the observations change; indeed, a $DNNF$ sentence $DNNF(SD \ | \ OBS)$
equivalent to the conditioning of $SD$ by the observations $OBS$ can
be computed as $DNNF(SD) \ | \ OBS$, the conditioning of a $DNNF$ sentence
equivalent to $SD$ by $OBS$. Since conditioning can be achieved
in linear time from a $DNNF$ formula, it is sufficient to compile
only the system description $SD$ (that is the fixed part of the diagnostic problem)
so to compute $DNNF(SD)$ instead of $DNNF(SD \ | \ OBS)$.

Because (1) forgetting variables in a $DNNF$ formula can be done
in polynomial time \cite{Darwiche99} and (2) the models of a
smooth $DNNF$ formula can be generated in time polynomial
in the output size \cite{darwicheJANCL}, we obtain that:

\begin{corollary} The most probable diagnoses for a diagnostic problem
${\cal P}$ can be enumerated in time polynomial in the size of
${\cal P}_{DNNF}$.
\end{corollary}

\section{CONCLUSION}

In this paper, we have studied how existing knowledge compilation functions
can be used to improve inference from propositional weighted
bases. Both negative and positive results have been put forward.
On the one hand, we have shown that the inference problem
from a compiled weighted base is as
difficult as in the general case, when prime implicates, Horn cover
or renamable Horn cover
target classes are considered.
On the other hand, we have shown that this problem becomes tractable whenever
$DNNF$-compilations are used. Finally, we have sketched
how our results can be used in model-based diagnosis in order to compute
the most probable diagnoses of a system.

\subsubsection*{Acknowledgements}
The first author has been partially supported by NSF grant IIS-9988543 and
MURI grant N00014-00-1-0617.
The second author has been partly supported by the IUT de Lens, the Universit\'e
d'Artois, the R\'egion Nord/Pas-de-Calais under the TACT-TIC project, and by the
European Community FEDER Program.

\end{document}